\pdfoutput=1

\documentclass[11pt]{article}

\usepackage[final]{acl}
\usepackage{enumitem}
\usepackage{array}
\usepackage{titlesec}
\usepackage{multicol}
\usepackage{fancyhdr}
\usepackage{graphicx}
\usepackage{adjustbox}
\usepackage{longtable}
\usepackage{caption}
\usepackage{booktabs}
\usepackage{times}
\usepackage{latexsym}

\usepackage[T1]{fontenc}

\usepackage[utf8]{inputenc}

\usepackage{microtype}

\usepackage{inconsolata}

\usepackage{graphicx}

\usepackage{multirow} 
\usepackage{booktabs} 
\usepackage{enumitem}

\title{{\sc SportSQL}: An Interactive System for Real-Time Sports Reasoning and Visualization }

\author{Sebastian Martinez \\
  Arizona State University \\
  \texttt{sjmart28@asu.edu}}

\author{
    Sebastian Martinez \quad Naman Ahuja \quad Fenil Bardoliya \quad Chris Bryan \quad Vivek Gupta \\
    Arizona State University \\
    \texttt{\{sjmart28, nahuja11, fbardoli, cbryan16, vgupt140\}@asu.edu}
}

\begin{document}
\maketitle
\begin{abstract}

We present a modular, interactive system \textsc{SportSQL} for natural language querying and visualization of dynamic sports data, with a focus on the English Premier League (EPL). The system translates user questions into executable SQL over a live, temporally indexed database constructed from real-time Fantasy Premier League (FPL) data. It supports both tabular and visual outputs, leveraging symbolic reasoning capabilities of Large Language Models (LLMs) for query parsing, schema linking, and visualization selection. To evaluate system performance, we introduce the \textbf{D}ynamic \textbf{S}port \textbf{Q}uestion \textbf{A}nswering benchmark (\textsc{DSQABench}), comprising 1,700+ queries annotated with SQL programs, gold answers, and database snapshots. Our demo highlights how non-expert users can seamlessly explore evolving sports statistics through a natural, conversational interface.

\end{abstract}

\section{Introduction}
What if a soccer fan could ask, \emph{“How did Mohamed Salah's scoring performance trend over the last five seasons?”} or \emph{“Which midfielders in the Premier League are the most creative this season?”} and instantly receive not only a precise answer but also a dynamic visualization, grounded in up-to-date, real-world data?

Large language models (LLMs) have shown remarkable progress in translating natural language into executable programs, such as SQL. However, most existing systems are designed for static, domain-specific datasets. In contrast, domains like sports are inherently dynamic and structurally complex: match outcomes, player statistics, team formations, and injury reports evolve daily across multiple interlinked and semi-structured tables. Querying such data effectively requires compositional, temporal, and relational reasoning, along with the ability to operate over continuously changing schemas and distributed sources.

We introduce \textsc{SportSQL}, a fully automated system for Dynamic Sports Question Answering (DSQA), enabling users to pose rich natural language queries over live sports data and receive grounded, executable, and often visual responses. \textsc{SportSQL} operates through a modular pipeline: it begins by scraping and normalizing dynamic data from transforming it into a unified, temporally indexed relational database. Given a user question, the system uses only the schema (not the data itself) to prompt an LLM to generate symbolic SQL queries, making the approach scalable and robust to changes in content \cite{kulkarni2025llm}. When appropriate, \textsc{SportSQL} also generates visualization code (in matplotlib, seaborn) to produce bar charts, timelines, or other graphical responses.

For instance, a user might ask, \textit{“Compare Arsenal’s goals scored in home vs away matches”} or \textit{“List forwards with at least ten goals and five assists.”} \textsc{SportSQL} retrieves accurate answers by executing SQL over the latest data, rather than relying on potentially outdated or hallucinated information from pretrained language models \cite{kulkarni2025reinforcing}. To evaluate the effectiveness of the system, we introduce Dynamic Sports Question Answering Benchmark (\textsc{DSQABench}), a new benchmark containing over 1700 questions that span various soccer metrics, reasoning types, and output formats. Each question is paired with its corresponding SQL program, gold answer, and the database snapshot at the time of execution. We further provide a type-aware evaluation framework that supports multiple answer formats, schema-only SQL generation, and fine-grained error analysis to assess system performance under dynamic conditions. Our contributions are threefold:


\begin{itemize}[itemsep=1pt]
\vspace{-0.25em}
    \item We introduce the task of Dynamic Sports Question Answering and present \textsc{SportSQL}, a modular and interpretable system that enables real-time, schema-driven symbolic reasoning and dynamic visualization over evolving sports databases.
    \item We construct and release \textsc{DSQABench}, the first benchmark of executable sports queries paired with live data, supporting multiple answer modalities.
    \item We develop a type-aware evaluation framework with support for diverse answer formats (textual, numeric, tabular, visual), schema-only SQL generation, and fine-grained error analysis to assess symbolic QA systems over dynamic content.
\end{itemize}

We invite the readers to explore SportsSQL's functionalities at the following links:

\begin{itemize}[itemsep=0.5pt, topsep=0.5pt, parsep=0.5pt, partopsep=0.5pt]
    \item Code $\&$ Data: \url{https://github.com/coral-lab-asu/SportSQL}
    \item Main Demo Video: \url{https://youtu.be/xqUyiA-R6aI}
    \item Longer Demo Video: \url{https://www.youtube.com/watch?v=3LnkkACfMmg}
    \item Try it out: \url{https://coral-lab-asu.github.io/SportsSQL}
\end{itemize}

Although \textsc{SportSQL} is designed for sports, its architecture is general and can extend to other dynamic, structured domains such as finance, healthcare, or elections, where users seek timely, accurate insights from evolving data.

\begin{figure*}[h]
  \centering
  \includegraphics[width=0.95\linewidth]{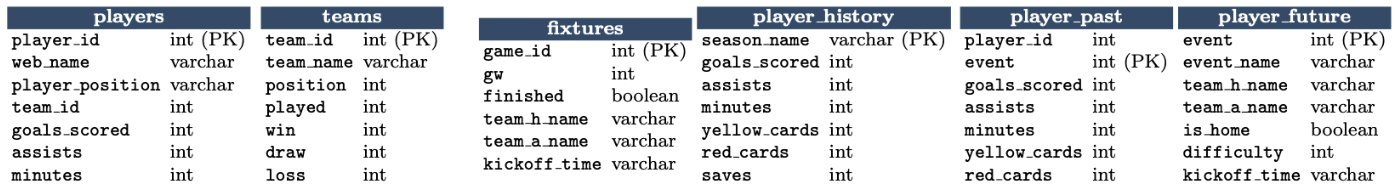}
  \vspace{-0.75em}
  \caption{DB schema, all tables shown, not all columns, Here, PK represent primary key.}
  \vspace{-0.75em}
  \label{fig:dbschema}
\end{figure*}

\begin{figure*}[h]
  \centering
  \includegraphics[width=0.95\linewidth]{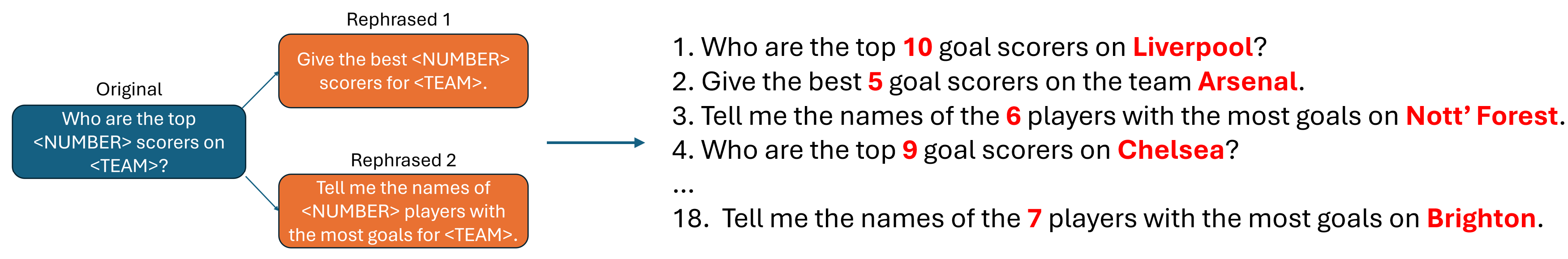}
  \vspace{-0.75em}
  \caption{Sample Question Creation Expansion}
  \label{fig:questioncreation}
  \vspace{-1.5em}
\end{figure*}

\section{\textsc{SportSQL} Architecture}
\textsc{SportSQL} translates free-form user queries into executable answers via a tightly integrated, multistage pipeline. The system operates over a live, dynamically updated EPL database, refreshed periodically via cronjobs and at runtime based on query requirements.
Upon receiving a natural language query, the system first performs entity grounding by executing SQL lookups against curated reference tables (e.g., teams, players), mapping surface forms to canonical entities. Conditioned on this context and the database schema, it generates an executable SQL query, which is run on the live database to produce a structured result.
If the query involves visual reasoning (e.g., comparisons, trends, rankings), the output is forwarded to a visualization agent, which selects an appropriate chart type and returns self-contained Python code (Matplotlib + Seaborn) to render the plot. The full workflow is outlined below.


\paragraph{1. Database Streaming}
Our system ingests data from the public \textit{Fantasy Premier League} (FPL) API,\footnote{\url{https://fantasy.premierleague.com/}} which offers structured, frequently updated endpoints covering players, teams, fixtures, and per-match statistics. After normalization and de-duplication, the data is stored in a MariaDB backend.
\paragraph{Hybrid Storage Strategy}
Storing full historical data for every player would require $\sim$2,400 tables per season (3 per player $\times$ 800 players), resulting in a bloated schema and largely idle data. To balance granularity with efficiency, we adopt a two-tiered storage design:
\begin{itemize}[itemsep=0.5pt]
\vspace{-0.25em}
\item \textbf{Query-agnostic tables}: Core relations (\texttt{players}, \texttt{teams}, \texttt{fixtures}) that evolve predictably week-by-week. These are updated nightly via cronjob to maintain freshness.
\item \textbf{Query-dependent tables}: Fine-grained views (e.g., “past 5 games”, “next 3 fixtures”) fetched on demand from the FPL API. These are materialized in memory for the duration of a query and discarded after use.
\vspace{-0.25em}
\end{itemize}
This hybrid architecture ensures (i) \textit{freshness} via automated updates, (ii) \textit{coverage} through just-in-time API access, and (iii) \textit{efficiency} by limiting persistent storage. Figure~\ref{fig:dbschema} illustrates the relational schema and data flow.

\paragraph{2. Entity Recognition}
User queries often contain abbreviations, nicknames, or informal spellings (e.g., \emph{“CR7”} for \emph{Cristiano Ronaldo}, \emph{“Donatello”} for \emph{Kylian Mbappé}), making exact string matching unreliable. Additionally, the LLM operates only over the database schema and lacks direct access to cell-level values.
To resolve entity mentions, we employ a prompt-guided procedure. The prompt instructs the LLM to:
(i) use domain knowledge to infer canonical player or team names, and
(ii) generate a case-insensitive wildcard SQL query over reference tables.
The database returns a filtered set of candidate rows with unique IDs, which are retained as the resolved entity identifiers.


\paragraph{3. SQL Generation and Execution}
Given the resolved entity identifiers, we prompt a large language model to generate an executable SQL query. The model is provided with:
(i) the user question,
(ii) the set of resolved primary keys, and
(iii) the database schema, along with targeted instructions to mitigate common pitfalls:
\begin{itemize}\setlength\itemsep{0em}
\item \textbf{Table hints}: e.g., \texttt{players} is preferred for individual statistics
\item \textbf{Synonym mappings}: e.g., “team position” $\leftrightarrow$ \texttt{league\_rank}
\item \textbf{Column cautions}: e.g., penalty saves are almost always non-zero for goalkeepers only
\item \textbf{Derived-field formulas}: e.g., \texttt{form} is the 30-day average of match points
\item \textbf{Scale explanations}: e.g., \texttt{strength} ranges from 1 (weakest) to 5 (strongest)
\end{itemize}
These prompt elements help ensure syntactic correctness and reduce semantic errors arising from natural language variability.


\vspace{0.5em}
\textit{\bf SQL Execution.} The generated SQL is parsed and executed against the dynamic MariaDB store.
If the query references a non-materialized \emph{query-dependent} table (e.g., a player’s upcoming fixtures), the system issues a just-in-time API call to fetch the necessary data, loads it into an in-memory buffer, and re-executes the query.
The temporary table is discarded post-aggregation, ensuring the persistent database remains lightweight.

\paragraph{4. Visual Output Generation}
Some information needs are better served through visualizations than text. To support this, the system automatically generates plots when either:
(i) the user explicitly requests a “plot,” “graph,” or “trend,” or
(ii) the output dataframe exhibits structures—such as multi-season time series or long categorical rankings—that benefit from visual interpretation.
For example, the query \emph{“Plot a line graph of Kylian Mbappé’s goal totals over the past five seasons”} produces a line chart with seasons on the $x$-axis and goals on the $y$-axis, revealing temporal trends. Similarly, the query \emph{“Which five teams recorded the highest average possession in the 2024–25 campaign?”}—though not explicitly visual—triggers a horizontal bar chart ranking clubs by possession.

When comparative or temporal reasoning is detected, the result and original query are passed to a secondary code-generating LLM, which returns self-contained \texttt{Matplotlib} code (e.g., line plots for trends, bar charts for rankings). A validation layer ensures the dataframe referenced in the code matches the SQL output byte-for-byte; any mismatch triggers automatic re-querying. 

\begin{figure*}[h]
  \centering
  \includegraphics[width=0.9\linewidth]{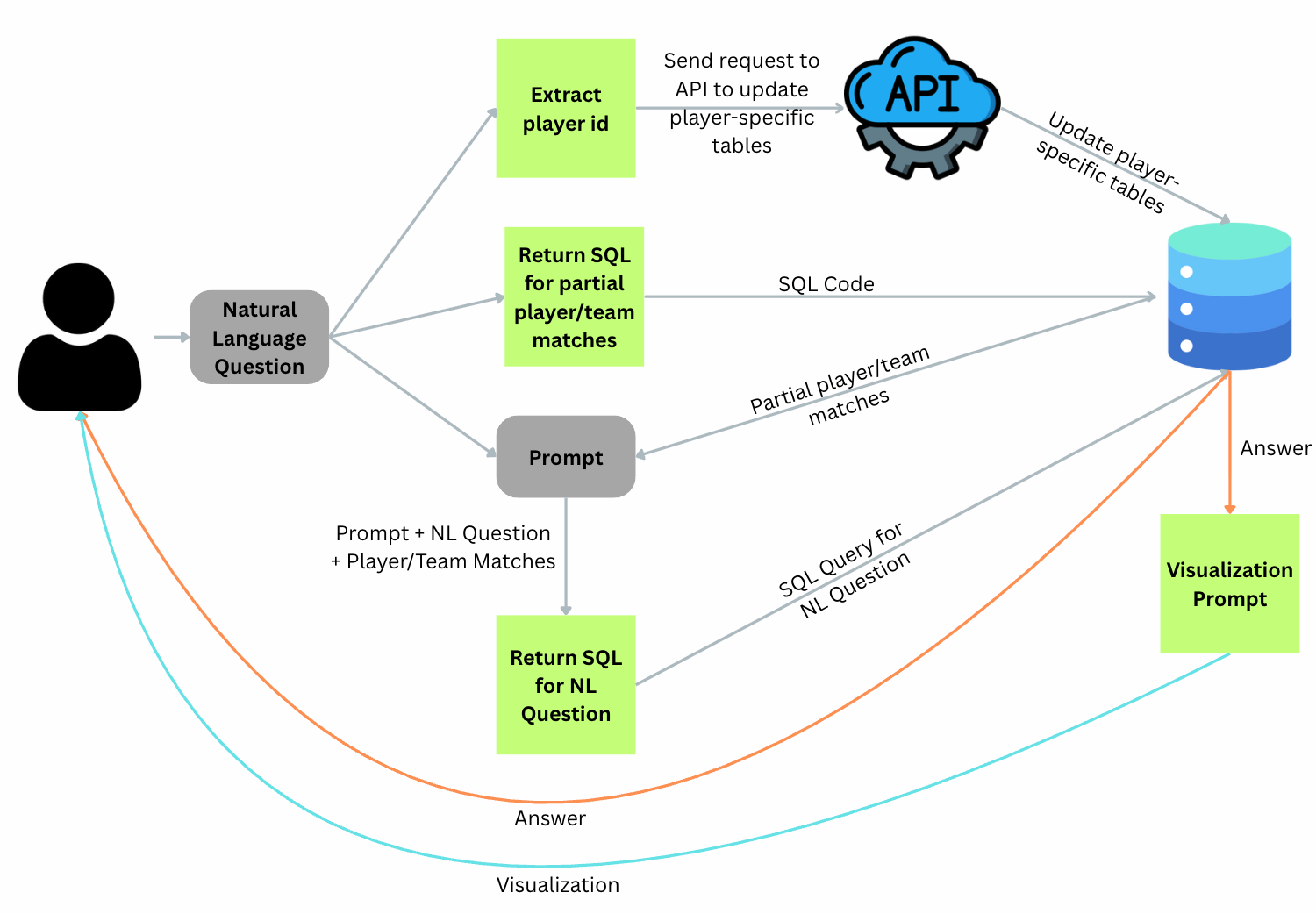}
  \vspace{-0.5em}
  \caption{SportSQL Architecture and Workflow}
  \label{fig:pipeline}
  \vspace{-0.75em}
\end{figure*}

This architecture enables near real-time visual responses, maintains the persistent database under 5GB, and supports fine-grained, player-level analytics without compromising freshness or correctness. Figure~\ref{fig:pipeline} presents an overview of the full system pipeline.

\section{\textsc{DSQABench} Benchmark}
To evaluate the \textsc{SportSQL} system, we introduce the \textbf{D}ynamic \textbf{S}port \textbf{Q}uestion \textbf{A}nswering benchmark (\textsc{DSQABench}), designed to assess natural language interfaces over dynamic, multi-relational sports data.

\paragraph{Query Creation.}
We construct a diverse set of natural language questions targeting various schema elements and reasoning skills. The process begins with manually written question templates, each rephrased to capture linguistic variation. Templates contain placeholders (e.g., team names, numerical thresholds), which are instantiated using real-world entities and context-appropriate values. This approach balances lexical diversity with semantic control. Additionally, we include a set of manually crafted, challenging questions to probe complex and multi-hop reasoning. An illustration of this process is shown in Figure~\ref{fig:questioncreation}.

\paragraph{Answer Annotation.}
Each question is paired with a manually authored SQL query, serving as the gold standard. These queries are executed against the underlying MariaDB-based \textit{``FutSQL\_FPL''} database to verify correctness. This ensures high-quality supervision for evaluating both SQL generation and execution accuracy.

\paragraph{Dataset Statistics.}
\textsc{DSQABench} contains 1,793 questions derived from 180 base templates, each rephrased in three distinct ways and instantiated with real-world values. Among these:
\begin{itemize}
\setlength\itemsep{0em}
\item 1,395 questions yield scalar answers (e.g., strings, numbers); 398 require tabular outputs.
\vspace{-0.5em}
\item 398 questions involve dynamic queries to player-specific tables via just-in-time API access:

\begin{itemize}[itemsep=0.0pt]
\item \texttt{player\_past}: 270 queries
\item \texttt{player\_history}: 72 queries
\item \texttt{player\_future}: 54 queries
\end{itemize}
\item All questions are paired with manually validated SQL programs executable on the database.
\end{itemize}
\textsc{DSQABench} provides a rich and realistic benchmark for studying compositional generalization, schema coverage, and executable reasoning in sports QA systems.

\section{Experiments and Analysis}

\paragraph{Models.}
We evaluate two state-of-the-art LLMs: \textsc{GPT-4o} and \textsc{Gemini-2.0 Flash}. \textsc{Gemini-2.0Flash} is selected for its balance of performance, latency, and cost, making it suitable for scalable deployment. \textsc{GPT-4o} is used to assess generalization. Both models use a temperature of 0.1 (for deterministic outputs) and a maximum token limit of 2048 (for reduced latency).
\paragraph{Evaluation Metrics.}
As the system produces both string and table outputs, we employ a type-aware evaluation. Rather than matching SQL queries, we directly compare outputs, as multiple queries may yield the same result.
\textbf{String Answers}: Evaluated using exact match.
\textbf{Table Answers}: Assessed using \textsc{TabEval}~\cite{ramu-etal-2024-bad}, which converts tables into atomic natural language statements and computes pairwise entailment via \textsc{RoBERTa-MNLI}, yielding precision (Correctness), recall (Completeness), and F1 (Overall) scores.



\subsection{Results and Analysis}

Table~\ref{tab:model_performance} reports performance on both string and table-structured questions. The system achieves up to 80\% exact-match accuracy and 0.75 macro-F1, indicating strong performance on structured QA. \textsc{GPT-4o} consistently outperforms \textsc{Gemini-2.0~Flash}, with gains of 4.2 points in exact match and 0.05 in macro-F1. Completeness scores exceed correctness for both models, suggesting that relevant columns are more reliably identified than specific rows, a reflection of the higher complexity of row selection driven by SQL predicates.


\begin{table}[!h]
    \small
    \setlength{\tabcolsep}{2pt}
    \centering
    \caption{Model performance comparison on string and tables answered. Here, EM represent as Exact Match.}
    \vspace{-0.5em}
\label{tab:model_performance}
    \begin{tabular}{l|c|c|c|c}
        \hline
        \multirow{2}{*}{\textbf{Model}} & \textbf{String} & \multicolumn{3}{c}{\textbf{Table (TabEval)}} \\
        \cline{2-5}
        & \textbf{EM} & \textbf{Correctness} & \textbf{Completeness} & \textbf{Overall} \\
        \hline
        Gemini-2.0 & 76.23 & 0.64 & 0.76 & 0.69 \\
        GPT-4o & 80.48 & 0.70 & 0.81 & 0.75 \\
        \hline
    \end{tabular}
\vspace{-1.0em}
\end{table}

\subsection{Primitive-Based Analysis}

To systematically assess performance across SQL query types, we annotate each ground-truth SQL template with a set of six reasoning primitives:
\begin{itemize}\setlength\itemsep{0em}
\vspace{-0.25em}
    \item \textbf{Calculate}: Arithmetic operations (\texttt{SUM}, \texttt{COUNT}, \texttt{AVG}, etc.)
    \vspace{-0.25em}
    \item \textbf{Compare}: Value comparisons
    \vspace{-0.25em}
    \item \textbf{Filter}: Conditional constraints 
    (\texttt{WHERE})
    \vspace{-0.25em}
    \item \textbf{Order}: Sorting (\texttt{ORDER BY ASC/DESC})
    \vspace{-0.25em}
    \item \textbf{Manipulate}: Data transformations (\texttt{JOIN}, \texttt{UNION}, \texttt{MERGE})
    \vspace{-0.25em}
    \item \textbf{Retrieve}: Direct lookups of values (e.g., entity or attribute selection)
\end{itemize}



\paragraph{Clause Combinations and Their Impact.}
The system performs exceptionally on single-primitive queries such as \emph{Retrieve} (\textit{“Show all EPL goalkeepers”}, 100\%) and \emph{Order} (\textit{“Rank Premier League clubs by points”}, 97.6\%). It also handles \emph{Calculate + Compare} well (\textit{“Did Haaland score more goals than Salah last season?”}, 96.3\%).
However, performance drops sharply with added complexity:
\emph{Retrieve + Filter + Calculate} (\textit{“What’s the average pass accuracy for midfielders under 23?”}) yields  22.3\%, and \emph{Compare + Order} reaches 30.3\%.
The most challenging cases involve \emph{Manipulate} operations (e.g., table joins) scoring 15.4\%, and four-way compositions (e.g., \emph{Compare + Manipulate + Order + Calculate}) showing similar results.

\begin{figure}[!htbp]
  \centering
  \includegraphics[width=0.9\linewidth]{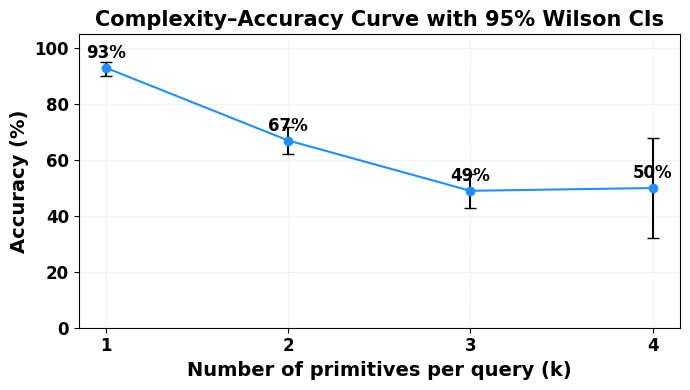}
  \vspace{-1.0em}
  \caption{Accuracy Trend over Number of Clauses}
  \label{fig:complexity}
  \vspace{-1.0em}
\end{figure}

\paragraph{Impact of Query Complexity.}
We examine how performance varies with the number of reasoning primitives in a query. Figure~\ref{fig:complexity} plots accuracy against the number of primitives ($k$), with 95\% Wilson confidence intervals.

Accuracy is high for single-step queries (93\%) but drops to 67\% with two primitives. Beyond three, performance stabilizes near 50\%, while confidence intervals widen, reflecting increased variance and data sparsity. This trend highlights the need to first improve dual-clause queries (e.g., \emph{Filter + Calculate}) before scaling to deeper compositions. It also suggests that future benchmarks should include more 3 and 4-primitive questions to better probe system limitations.

\begin{figure}[!htbp]
\vspace{-0.5em}
  \centering
\includegraphics[width=0.95\linewidth]{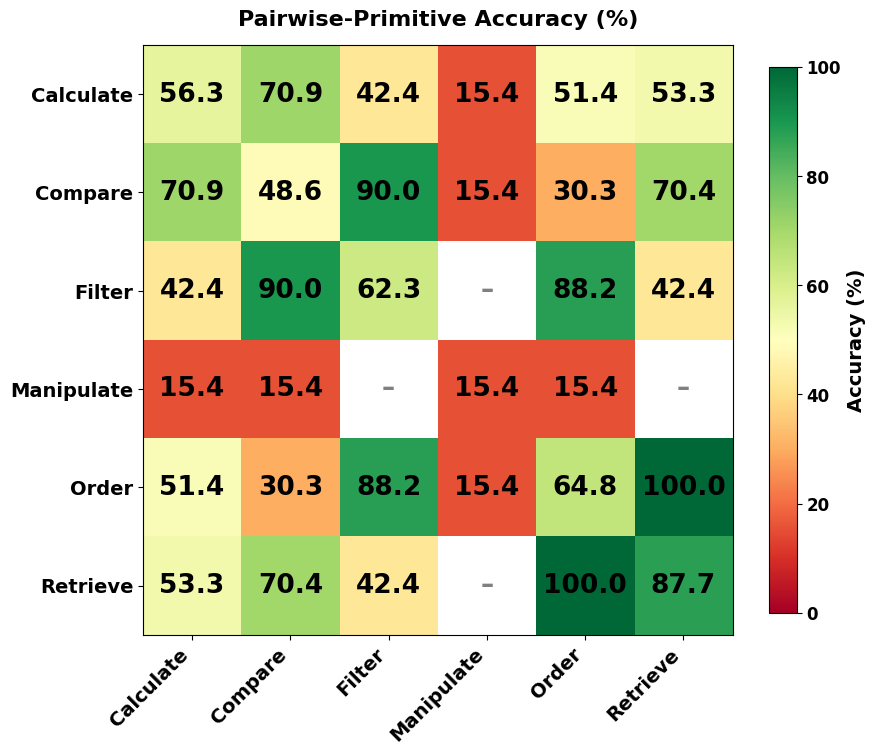}
  \vspace{-1.0em}
  \caption{Pairwise Primitive Accuracy}
  \label{fig:heatmap}
  \vspace{-1.0em}
\end{figure}

\paragraph{Bottleneck Clause Pairs.}
Figure~\ref{fig:heatmap} shows accuracy for all pairs of reasoning primitives.
\emph{Retrieve + Order} performs well ($\geq$88\%), indicating strong compatibility between basic operations.
In contrast, any pair involving \emph{Manipulate} or \emph{Calculate} drops sharply to around 15\%, even when combined with otherwise reliable primitives like \emph{Compare}.
These patterns align with the decline in Figure~\ref{fig:complexity}, where queries involving aggregation or table restructuring introduce significant error.
Overall, the heatmap identifies \textbf{aggregation and manipulation} as key bottlenecks for improvement.


\section{Qualitative Example}
\vspace{-0.5em}

\vspace{0.5em}
\noindent\textbf{Query:} \textit{Show me the top 10 goal scorers and their goal count.}

\vspace{0.25em}
\noindent \textbf{Generated SQL:}
\vspace{-0.5em}
\begin{verbatim}
SELECT web_name, goals_scored 
    FROM players 
    ORDER BY goals_scored DESC 
    LIMIT 10;
\end{verbatim}

\noindent\textbf{Generated Table:}
\begin{center}
\small
\begin{tabular}{l c}
\toprule
\textbf{web\_name} & \textbf{goals\_scored} \\
\midrule
M.Salah   & 27 \\
Haaland   & 21 \\
Isak      & 20 \\
Wood      & 18 \\
Mbeumo    & 16 \\
Watkins   & 14 \\
Wissa     & 14 \\
Palmer    & 14 \\
Cunha     & 14 \\
Mateta    & 13 \\
\bottomrule
\end{tabular}
\end{center}

\vspace{0.25em}
\noindent \textbf{Generated Visual:}
\vspace{-0.25em}
\begin{figure}[h]
  \centering
  \includegraphics[width=1\linewidth]{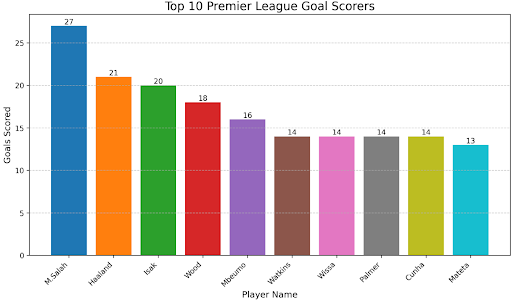}
  \label{fig:fig1}
  \vspace{-2.5em}
\end{figure}

\section{Related Works}

\paragraph{Text-to-SQL.}
Text-to-SQL research has primarily framed the task as cross-domain semantic parsing over static relational schemas. Benchmarks like Spider \cite{yu-etal-2018-spider} and its extensions \cite{li2023can, zhang2024benchmarking, pourreza-rafiei-2023-evaluating} focus on generalization to unseen databases, yet operate over fixed snapshots with limited domain dynamics. SyntaxSQLNet \cite{yu-etal-2018-syntaxsqlnet} introduced syntax-tree decoders for nested queries, while recent advances \cite{zhang-etal-2023-act, xie-etal-2024-decomposition} improve compositional reasoning and execution accuracy. 

However, these methods assume immutable schemas, overlook temporal drift in cell values, and sidestep challenges like domain-specific entity resolution (e.g., player aliases) that arise in continuously evolving datasets.

\paragraph{Sports QA.}
Prior work in sports question answering has largely centered on unstructured text or multiple-choice formats. LiveQA \cite{qianying-etal-2020-liveqa} explores NBA commentary, using timeline-based MCQs grounded in broadcast text. AskSport \cite{stoisser2025sparks} retrieves top-k passages via BM25+RoBERTa, but lacks symbolic execution and numerical guarantees. These systems do not support natural language aggregation (e.g., “average points in last 5 matches”) or multi-table joins—capabilities native to SQL. 

Our work bridges Text-to-SQL and SportsQA by introducing \textsc{SportSQL}, a pipeline tailored to dynamic sports data, and \textsc{DSQABench}, the first benchmark pairing natural language queries with executable SQL over temporally indexed, continuously refreshed soccer statistics.

\section{Conclusion and Future Work}
\textsc{SportSQL} demonstrates how natural language interfaces can make complex, evolving sports data accessible to everyday users without technical expertise. By combining structured prompt engineering with real-time data integration and multimodal outputs, the system offers a robust and extensible platform for interactive sports analytics. The release of \textsc{DSQABench} provides a valuable resource for benchmarking and advancing research in dynamic, temporally grounded question answering.

In future work, we plan to (1) support more advanced query types, including comparative and multi-turn analyses across players, teams, and seasons, and (2) generalize the framework to additional structured domains such as finance, healthcare, and other sports like basketball or American soccer. This work lays the groundwork for scalable, domain-agnostic natural language access to complex, real-world databases.

\section{Limitations}
\label{sec:limitations}

While our system performs well on natural language to SQL translation over dynamic sports data, several limitations remain. First, ranked queries using \texttt{LIMIT} (e.g., “top 5 goal scorers”) may omit tied results due to default lexicographic ordering, yielding incomplete answers. Second, the system supports only English input, limiting accessibility for multilingual users. Third, context length constraints restrict the ability to encode real-time metadata such as recent transfers or lineup changes.

Moreover, the current system is tailored to the English Premier League and does not readily generalize to other sports or leagues without domain-specific adaptation. Expanding to new domains would require schema remapping and possible model fine-tuning. Future work may incorporate multilingual LLMs, retrieval-augmented generation, and adaptive components to improve robustness across languages, domains, and evolving contexts.


\bibliography{anthology,custom}

\section{Appendix}
\label{sec:appendix}


\noindent\textbf{Query 1:} \textit{Give me the player history table for James Milner.}

\vspace{0.25em}
\noindent \textbf{Generated SQL 1:}
\vspace{-0.5em}
\begin{verbatim}
SELECT * FROM player_history;
\end{verbatim}

\vspace{-0.25em}
\noindent\textbf{Generated Table 1:}
\begin{center}
\footnotesize
\setlength{\tabcolsep}{1pt}
\begin{tabular}{l c c c c c}
\toprule
\textbf{season} & \textbf{total} & \textbf{minutes} & \textbf{goals} & \textbf{assists} & \textbf{clean} \\
\textbf{name} & \textbf{points} & & \bf scored & & \bf sheet \\
\midrule
2006/07  & 114 & 2675 & 3 & 7  & 0 \\
2007/08  & 84  & 2227 & 2 & 2  & 0 \\
2008/09  & 128 & 3060 & 3 & 9  & 0 \\
2009/10  & 184 & 3172 & 7 & 12 & 0 \\
2010/11  & 97  & 2134 & 1 & 7  & 11 \\
2011/12  & 86  & 1586 & 3 & 5  & 6 \\
2012/13  & 96  & 1724 & 4 & 4  & 11 \\
2013/14  & 67  & 1373 & 1 & 6  & 5 \\
2014/15  & 107 & 1749 & 5 & 8  & 7 \\
2015/16  & 123 & 2409 & 5 & 11 & 8 \\
2016/17  & 139 & 3154 & 7 & 4  & 12 \\
2017/18  & 77  & 1759 & 0 & 3  & 6 \\
2018/19  & 101 & 1778 & 5 & 5  & 9 \\
2019/20  & 49  & 924  & 2 & 2  & 4 \\
2020/21  & 44  & 1056 & 0 & 2  & 4 \\
2021/22  & 38  & 844  & 0 & 1  & 4 \\
2022/23  & 42  & 889  & 0 & 1  & 3 \\
2023/24  & 28  & 770  & 0 & 2  & 4 \\
\bottomrule
\end{tabular}
\end{center}

\begin{figure}[!htbp]
  \centering
  \includegraphics[width=1\linewidth]{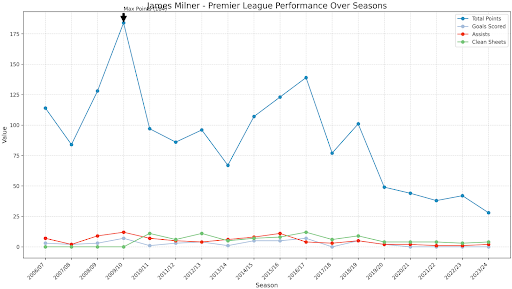}
  \caption{Visualization Output Query 1.}
  \label{fig:fig2}
\end{figure}



\noindent\textbf{Query 2:} \textit{Show me the team names, positions, points, and strength in a color scatterplot.}

\vspace{0.25em}
\noindent \textbf{Generated SQL 2:}
\vspace{-0.5em}
\begin{verbatim}
SELECT team_name, position, points, strength 
FROM teams;
\end{verbatim}

\vspace{-0.25em}
\noindent\textbf{Generated Table 2:}
\begin{center}
\small
\begin{tabular}{l c c c}
\toprule
\textbf{team\_name} & \textbf{position} & \textbf{points} & \textbf{strength} \\
\midrule
Liverpool      & 1  & 76 & 5 \\
Arsenal        & 2  & 63 & 4 \\
Nott'm Forest  & 3  & 57 & 4 \\
Newcastle      & 4  & 56 & 4 \\
Man City       & 5  & 55 & 4 \\
Chelsea        & 6  & 54 & 4 \\
Aston Villa    & 7  & 54 & 3 \\
Bournemouth    & 8  & 48 & 4 \\
Fulham         & 9  & 48 & 3 \\
Brighton       & 10 & 48 & 3 \\
Brentford      & 11 & 43 & 3 \\
Crystal Palace & 12 & 43 & 3 \\
Everton        & 13 & 38 & 3 \\
Man Utd        & 14 & 38 & 3 \\
Spurs          & 15 & 37 & 3 \\
Wolves         & 16 & 35 & 3 \\
West Ham       & 17 & 35 & 3 \\
Ipswich        & 18 & 21 & 3 \\
Leicester      & 19 & 18 & 3 \\
Southampton    & 20 & 10 & 2 \\
\bottomrule
\end{tabular}
\end{center}

\begin{figure}[!htbp]
  \centering
  \includegraphics[width=1\linewidth]{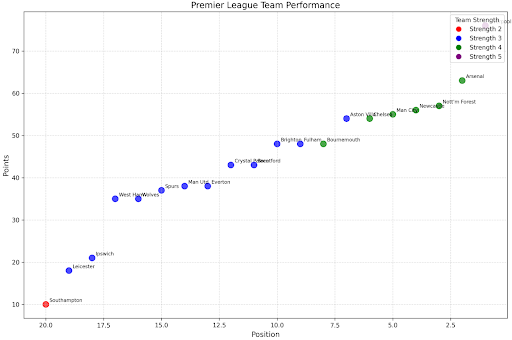}
  \caption{Visualization Output Query 2.}
  \label{fig:fig3}
\end{figure}

\begin{figure*}[h]
  \centering
  \includegraphics[width=1\linewidth]{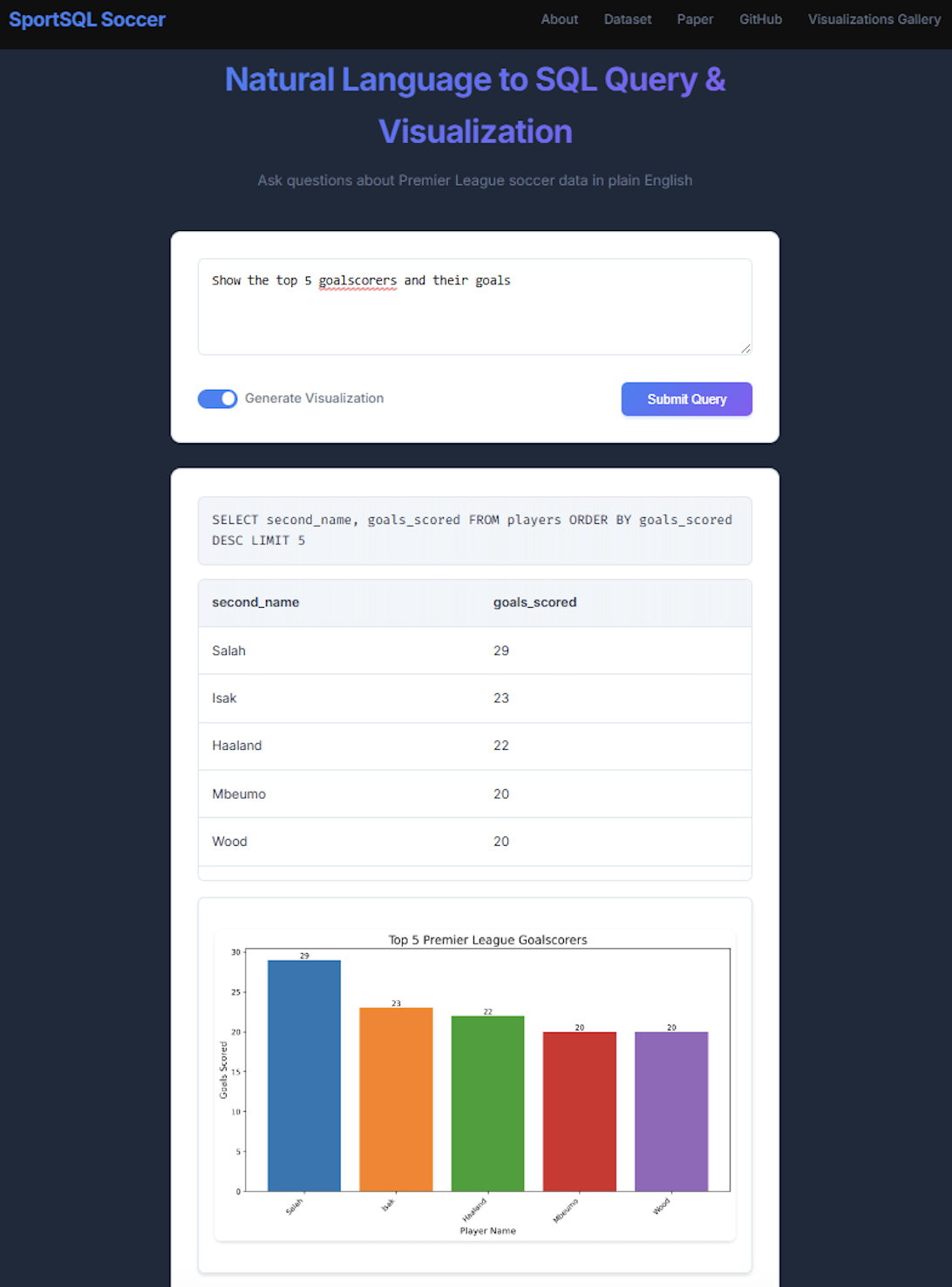}
  \caption{SportSQL System Demonstration.}
  \label{fig:fulldemo}
\end{figure*}

\end{document}